\documentclass[final,5p,times,twocolumn]{elsarticle}

\usepackage{amssymb}
\usepackage{amsmath}

\journal{Preprint}

\usepackage{framed,multirow}
\usepackage{multirow}
\usepackage{diagbox}
\usepackage{subcaption}
\usepackage{amssymb}
\usepackage{amsmath}
\usepackage{latexsym}

\usepackage{url}
\usepackage{xcolor}
\definecolor{newcolor}{rgb}{.8,.349,.1}

\begin{document}

\begin{frontmatter}

\title{Performance of Gaussian Mixture Model Classifiers on Embedded Feature Spaces}

\author{Jeremy  Chopin} 
\author{Rozenn  Dahyot\corref{cor1}}

\cortext[cor1]{Corresponding author: Rozenn Dahyot
\ead{Rozenn.Dahyot@mu.ie}
}

\affiliation{organization={Computer Science Department},
                addressline={Maynooth University}, 
                city={Maynooth}, 
                country={Ireland}}

\begin{abstract}
Data embeddings with CLIP \citep{CLIP2021} and ImageBind \citep{Girdhar_2023_CVPR} provide powerful features for the analysis of  multimedia and/or multimodal data.
We assess their performance here for classification using  a Gaussian Mixture models (GMMs) based layer as an alternative to the standard Softmax layer.
 GMMs based classifiers  have recently been shown  to have interesting performances as part of deep learning pipelines  trained end-to-end \citep{hayashi2021a}. 
Our first contribution is to investigate GMM based classification performance taking advantage of the embedded spaces CLIP \citep{CLIP2021} and ImageBind \citep{Girdhar_2023_CVPR}.
Our second contribution is in proposing our own GMM based classifier with a lower parameters count than previously proposed \citep{hayashi2021a}. 
Our findings are, that in most cases, on these tested embedded spaces, one gaussian component in the GMMs is often  enough for capturing each class,
and  we hypothesize that this may be due to  the contrastive loss used for training these embedded spaces that naturally concentrates features together  for each class. We also observed that   ImageBind \citep{Girdhar_2023_CVPR} often provides better performance than CLIP \citep{CLIP2021} for classification of image datasets
even when these embedded spaces are compressed using PCA.  
\end{abstract}


\end{frontmatter}

\section{Introduction}

Gaussian mixture models (GMMs) are well known  powerful parametric probability density functions,
popular for modelling likelihoods for classes of  interest in $D$ dimensional feature spaces. 
These likelihood-GMMs  together with their respective class priors can then be used for defining a Bayes Classifier.  
Hayashi and Uchida \cite{hayashi2021a} have proposed  the Sparse Discriminative Gaussian mixture (SDGM) layer 
and have compared it with Softmax classifier as part of several deep learning pipelines that are trained end-to-end  with the classifier layer. 
Here we first evaluate SDGM  on  data embeddings computed  with CLIP \citep{CLIP2021} and ImageBind \citep{Girdhar_2023_CVPR}.

 SDGM \cite{hayashi2021a} estimates  GMMs with the choice of full covariance matrices (with $D(D-1)/2$ parameters for each covariance; the resulting classifier is noted SDGM-F) 
 and  lower dimensional diagonal covariance matrices  (with $D$ parameters for each covariance; the resulting classifier is noted SDGM-D). 
 We propose our own  deep neural network layer  called  Deep Gaussian Mixture Model Classifier (DGMMC)
 that similarly is able to fit both  GMMs as likelihoods and estimate their corresponding  class priors, for  performing  
Bayesian classification.  DGMMC\footnote{DGMMC code provided at \url{https://github.com/CVMLmu/DGMMC/}.} 
has a different implementation to SDGM\footnote{SDGM \url{https://github.com/HideakiHayashi/SDGM_ICLR2021/}.}
and it  also allows to use spherical covariance matrices (noted DGMMC-S) 
that has a much lower number of parameters (only 1 parameter per covariance matrices).  
Our new layer for classification DGMMC is presented in Section \ref{sec:DGMMC} and its complexity is compared to SDGM in Section \ref{sec:parameters:DGMMC:SDGM}.

As part of a classifier layer, a linear operation   can be learnt for reducing (or extending) the dimensionality $D$ of the embedding space before fitting GMMs.
We explore several strategies in Section \ref{sec:classification:results} for that linear operation 
when we benchmark our approach for classification on several image datasets to demonstrate DGMMC  performance.  
Overall our findings are that accuracies are better on CLIP and ImageBind than with reported end to end pipelines \cite{hayashi2021a}, 
that ImageBind  performs better than CLIP, and the number of Gaussians required in the GMMs can  be kept small (e.g $G=1$) 
on these  embedded spaces. A summary table of our accuracy results is shown Tab. \ref{tab:Summary}.
We start next  by providing an overview of the relevant literature.

\begin{table}[!h]
\begin{tabular}{ccccc}
MNIST & CIFAR10&CIFAR100&ImageNet&ESC50\\
\hline
\hline
\textbf{99.3} &	\textbf{98.8}	&\textbf{91.2}&	\textbf{84.1}&	\textbf{87}\\
\footnotesize{DGMMC-S}	&
\footnotesize{DGMMC-S}&	
\footnotesize{DGMMC-S}	&
\footnotesize{DGMMC-S}	&
\footnotesize{SDGM-D}\\
 Tab.\ref{tab:pretrained_results}    &  Tab. \ref{tab:pretrained_CIFAR100}    &    Tab.\ref{tab:pretrained_results}   &  
  Tab.\ref{tab:pretrained_results}& Tab.\ref{tab:pretrained_ESC50}\\
\hline
\hline
$99.14^{\star}$ &	$89.54^{\star}$ &	$78.68^{\star}$ &	$67.6^{\star}$&	$66.9^{\circ}$\\
\hline
\end{tabular}
\caption{For the tested datasets (top row), \textbf{our best classification accuracies obtained with ImageBind} are reported  (second row) with the names of the classifier  (3rd row).  The corresponding Tables where these results are extracted from, are indicated in the 4th row. All models are with $G=1$ except for CIFAR10 and ESC50 where $G=2$. For comparison, the reported best accuracies from \cite{hayashi2021a}$^{\star}$ (with an end-to-end pipeline using SDGM-F) and \cite{Girdhar_2023_CVPR}$^{\circ}$ (zero shot classification) are shown on the bottom line.}
\label{tab:Summary}
\end{table}

\section{Related works}
\label{sec:SOTA}

\subsection{Classification of images}

Convolutional Neural Networks (CNNs) have demonstrated excellent performance for image classification  and some recent efforts have focused on efficient computing of convolution layers  \cite{ULICNY2022108707} or learning activation functions that follows convolution operations \cite{DeepSplines2020}. 
More recently Transformers (ViTs) \cite{dosovitskiy2021an} have been proposed as an alternative  to CNNs, that are able to take advantage of long range spatial correlation in images that are often missed when using CNNs. 
Currently, Softmax is used as the final layer in both CNNs and  Vision Transformers (ViTs) for classification decision \cite{DeepSplines2020,dosovitskiy2021an,ULICNY2022108707}.

\subsection{Classifiers and GMMs}

Classifiers that are based on a Softmax layer have achieved remarkable performance for the image classification task. 
However, as summarized in the work of  Liang et al \cite{Chen2022}, these classifiers have shortcomings. 
First, the decision boundaries are learnt between the $\mathcal{C}$ classes in the $D$-dimensional feature space but does not learn the intrinsic class characteristics and can be limited for generalizing on unseen data. 
Secondly, each class $c$ is captured by a single projection shifted with a bias hence the implicit assumption of the unimodality of each class in the feature space. Finally, the prediction score of a class is unreliable for uncertainty modelling and is only used  for its comparative value against other classes.
classifiers can focus instead on learning the joint data distribution using for instance Gaussian Mixture models (GMMs)  \cite{Chen2022}.  
Indeed GMMs are  powerful probabilistic models that  estimate any density provided there is enough Gaussian components in the mixture \cite{Jian2011}. 
GMMs have been used for  image classification. For instance  Wan et al ~\cite{Wan2019} train their GMMs using the Expectation-Maximization (EM) algorithm with an additional criteria to choose the adequate number of Gaussian components in the GMMs.  GMMs have also been used in the context of deep learning for  image classification: for instance Celarek et al~\cite{celarek2022}  propose  a convolutional neural network layer processing a GMM (input) into another GMM (output)  with  learnable filters also defined as GMMs. 
However their architecture uses Softmax for generating class scores \cite{celarek2022}.
Zhang et al.~\cite{zhang2021} used a Gaussian mixture as a measure of the attention in Transformer Model for Natural Language Processing (NLP) tasks. Gepperth et al ~\cite{Gepperth2021} proposed to fit GMMs using a Stochastic Gradient Descent instead of the EM algorithm and obtained remarkable results, encouraging us to include the estimation of the parameters of the GMMs as parts of the neural network model.

Liang et al \cite{Chen2022} recently proposed the use of GMMs as a  classifier and obtain  competitive results for  semantic segmentation on benchmark datasets. Their  model is composed of a feature extractor and a classifier based on GMMs trained with an hybrid training, where the feature extractor is trained using a Softmax layer (discrimative classifier) and the GMMs of the classifier are trained using a modified EM Algorithm based on optimal transport \cite{Cuturi2013}. 

Closely related to our work is the sparse discriminative Gaussian mixture (SDGM) layer \cite{hayashi2021a} that has been proposed for classification and compared favorably with traditional classification techniques and Softmax layer on synthetic and image datasets.   
In Section \ref{sec:DGMMC}, we propose likewise a   classifier layer using GMMs where all  parameters of the Neural Networks and  of GMM classifier layer (i.e. means, covariance matrix and weigths)  can be optimized end-to-end using standard optimisation used with deep neural networks.

\subsection{Embedding spaces with contrastive learning}

Contrastive learning aims at learning a representation such that certain pairs of inputs (positive pairs),
are close in the embedding space, while other pairs of inputs (negative pairs) are far apart 
\cite{Bishop2024}.
In contrastive-language image pretraining (CLIP) \cite{CLIP2021}, a positive pair consists of an image and its corresponding
text caption, that are processed with two separate functions (one for each modality) and these  positive pairs  are close inputs to the embedding space. 
ImageBind \cite{Girdhar_2023_CVPR} extends this idea using positive pairs of an image with additional modalities including for instance text (like CLIP), audio and other type of images (e.g. depth, thermal). 
The images in the pairs $\lbrace$(image, other data)$\rbrace$  is sufficient to bind the data modalities together \cite{Girdhar_2023_CVPR}.
Separate transformers are used to process each modalities,  for instance  ViTs \cite{dosovitskiy2021an} are used to process image modalities for both (CLIP) \cite{CLIP2021} and ImageBind \cite{Girdhar_2023_CVPR}.
In this paper we have used  the two pre-trained transformers  provided by CLIP and ImageBind  for processing image datasets, hence creating powerful features for our classifiers. In addition we also show classification results on an audio dataset  that has been processed with the ImageBind pre-trained audio function
(cf. Sec. \ref{sec:classification:results}).

\section{DGMMC: Deep Gaussian Mixture Model Classifier}
\label{sec:DGMMC}

In the following, we denote $x$ a feature (e.g. computed with CLIP or ImageBind) to classify into $\mathcal{C}$ classes. The likelihood $p(x|c)$ for class $c=1,\cdots,\mathcal{C}$ is modelled as a GMM with $G$ components (Gaussians).

\subsection{GMMs: Gaussian Mixture Models}

The posterior probabilities $\lbrace p(c|x) \rbrace_{c=1,\cdots,\mathcal{C} }$ are represented with Gaussian Mixture Models, and features extracted from  images are used as  inputs for our GMMs classifier. The goal here is to model the joint distribution $p(x, c)$ by estimating both the class-specific distribution $p(x|c)$ and the prior probability of each class $p(c)$. Then using Bayes rules we can estimate the posterior probabilities as:
\begin{equation}
\label{bayes}
    p(c|x) = \frac{p(x|c)\ p(c)}{\sum^{\mathcal{C}}_{c'=1} p(x|c')\ p(c') }
\end{equation}
where $\mathcal{C}$ is the number of classes considered.
The GMM $p(x|c)$  is composed of multivariate Gaussians for embedded features $x\in\mathbb{R}^D$ noted:
\begin{equation}
\label{pdf_GMM}
    p(x|c) = \sum^{k^c}_{i=1}\omega^c_i\ \phi(x|\mu^c_i, \Sigma^c_i)
\end{equation}
 where $k^c$ is the number of Gaussian components in the mixture for the class $c$. $\phi$ is the multivariate Gaussian distribution and $\omega^c_i$, $\mu^c_i$, $\Sigma^c_i$ are respectively the weights, the means and the covariance matrix of the Gaussian component $i$ in the mixture for the class $c$. 
 To limit the complexity of estimating $D\times D$ covariance matrices $\lbrace \Sigma^c_i \rbrace$ having $\frac{D\times (D+1)}{2}$ parameters, spherical Gaussian mixtures are considered instead \cite{Gepperth2021}:
where $ \Sigma^c_i = b^c_i\ \mathrm{I}_D$ with bandwidth $b^c_i$ and identity matrix $\mathrm{I}_D$. In this case, Equation \ref{bayes} becomes:
\begin{equation}
\label{bayes_gmm}
    p(c|x) = \frac{p(c) \sum^{G}_{i=1} \omega_{i}^c \ \phi(x|\mu_{i}^c, b^c_i\ \mathrm{I}_D)}{\sum^{C}_{c'=1} \bigr[ p(c')\sum^{G}_{i=1}\omega^{c'}_i\ \phi(x|\mu^{c'}_i, b^{c'}_i\ \mathrm{I}_D) \bigr]}
\end{equation}
with the additional constraint for that all GMMs have the same number of components: $k^c=G, \forall c\in\lbrace 1,\cdots,\mathcal{C}\rbrace$.

\subsection{Deep Gaussian Mixture Model}
\label{sec:DGMM}

Our classifier is trained as part of a deep neural network using Stochastic Gradient Descent (SGD). 
Based on equation \ref{bayes_gmm}, the following tensors are defined as parameters to estimate in our layer:  
\begin{itemize}
\item $\mathrm{P} \in \mathbb{R}^{\mathcal{C}}$ stores the prior $p(c)$ of each class and it  has the following constraint: 
\begin{equation}
\left(\sum_{c=1}^{\mathcal{C}} P_c= 1 \right) \quad \wedge \quad\left( P_c\geq 0, \ \forall c=1,\cdots,\mathcal{C}\right)
\end{equation}
\item $\mathrm{W} \in \mathbb{R}^{\mathcal{C} \times G}$ is capturing the positive weights of the Gaussian components for all $\mathcal{C}$ classes GMMs, and has the following constraint: 
\begin{equation}
\left(\forall c \in \mathcal{C}, \ \mathrm{W} \mathbf{1}_{G} = \mathbf{1}_{\mathcal{C}}\right) \quad \wedge   \quad\left( W_{i,j}\geq 0,  \ \forall (i,j) \right)
\end{equation}
with the notation $\mathbf{1}_{G}$ (resp. $\mathbf{1}_{\mathcal{C}}$ ) for a vector of ones in $\mathbb{R}^G$ (resp. in $\mathbb{R}^{\mathcal{C}}$).
\item $\mathrm{M} \in \mathbb{R}^{\mathcal{C} \times G \times D}$ is the tensor collecting all  the mean vectors in $\mathbb{R}^D$. 

\item Finally, the tensor $\mathrm{B} \in \mathbb{R}^{\mathcal{C} \times G}$ encodes the positive bandwidths with the constraint $B_{i,j}>0, \ \forall (i,j)$.

\end{itemize}
The softmax operation is used to enforce the constraints on parameter tensors $\mathrm{P}$ and $\mathrm{W}$.
For the bandwidths $\mathrm{B}$, these are clamped with a minimal value set to be close to zero ($1e-6$).
Our classifier is trained using SGD optimizer \cite{Gepperth2021} with the cross-entropy loss.

\subsection{Computational complexity DGMMC}

Table \ref{tab:DGMMC:nb:parameter} summarizes the number of parameters in our proposed models DGMMCs.
The main gain that can be achieved is in choosing spherical  covariance matrices and consequently in this paper 
we focus on our low parameter model DGMMC-S to compare with SDGM-F and SDGM-D proposed by Hayashi and Uchida \cite{hayashi2021a}.  
The total number of parameters $P_{rm}$ to estimate in our classifier layer DGMMC-S can be calculated as follow: 
\begin{equation}
\label{eq:total_parameters}
    P_{rm} = \mathcal{C}\times (G\times (D+2) + 1)
\end{equation}
\begin{table}[!h]
\centering
\begin{tabular}{lcccc}
\hline
& $\lbrace\mu_i^c\rbrace$ & $\lbrace p(c)\rbrace$* & $\lbrace\omega_i^c\rbrace$** & $\lbrace\Sigma_i^c\rbrace$\\
\hline
\hline
DGMMC-F &$DGC$ & $C$ & $GC$ & $D^2CG$\\
DGMMC-D &$DGC$ & $C$ & $GC$ & $DCG$\\
\hline
DGMMC-S &$DGC$ & $C$ & $GC$ & $CG$\\
\hline
\end{tabular}
\caption{Number of learnt parameters for DGMMC classifiers with  *formula for the number of classes $C\geq 2$ (task of classification in at least 2 classes) and  **formula for the number of Gaussian components $G\geq 2$ (for $G=1$,  there is no learnt parameter as $\omega^c=1,\ \forall c$) in a feature space of dimension $D$. } \label{tab:DGMMC:nb:parameter}
\end{table}

\begin{table*}[!h]
    \centering
\begin{tabular}{cc|cc|cc|cccc|}
\cline{3-10} 
&& \multicolumn{2}{c|}{$C=10$} &\multicolumn{2}{c|}{$C=100$} & \multicolumn{4}{c|}{$C=1000$}\\

&  & $d=2$ & $d=10$& $d=32$ & $d=100$ & $d=128$ & $d=768$ & $d=1000$ & $d=1024$\\
 \hline
  \hline
\multirow{2}{*}{$\times$SDGM-F Eq. (\ref{eq:ratio:SDGM-F})} & G=1 & 2.75$\times$& 10.92$\times$ & 32.97$\times$ & 101$\times$ & 129$\times$& 769$\times$ & 1001$\times$ &	1025$\times$  \\
 & G=2 & 2.44$\times$ & 10.48$\times$ & 32.49$\times$ & 100.5$\times$ & 128.5$\times$& 768.5$\times$ & 1000.5$\times$& 1024.5$\times$ \\
 \hline
\multirow{2}{*}{$\times$SDGM-D Eq. (\ref{eq:ratio:SDGM-D})} & G=1 & 2.25$\times$ & 3.42$\times$ & 3.79$\times$& 3.93$\times$ & 3.95$\times$& 3.99$\times$ & 3.99$\times$ & 3.99$\times$ \\
 & G=2 & 2$\times$ & 3.28$\times$ & 3.74$\times$ & 3.91$\times$& 3.93$\times$& 3.99$\times$& 3.99$\times$& 3.99$\times$ \\
 \hline
 \hline
 \multirow{2}{*}{\# DGMMC-S}& G=1 & \#40 & \#120& \#3400 & \#10200 & \#130000& \#770000 & \#1002000& \#1026000  \\
 & G=2 & \#90 & \#250& \#6900 & \#20500 & \#261000& \#1541000& \#2005000 &  \#2053000\\
 \hline
\end{tabular}
\caption{Ratios of the number of parameters in SDGM-F and SDGM-D  w.r.t. our model DGMMC-S. The numbers of parameters in DGMMC-S are provided in the bottom row (see Sec. \ref{sec:parameters:DGMMC:SDGM}). The choices for dimension $d$, number of classes $C$ and number of Gaussian components $G=\lbrace 1,2 \rbrace$ correspond to the classification accuracy results reported in the next Section \ref{sec:classification:results}.}
\label{fig:model_parameters}
\end{table*}

\subsection{Remarks}
\label{sec:remarks}

We have used one GMM per class where each GMM is composed of a number of $G$ Gaussian components in the feature space of dimension $D$ as in \cite{hayashi2021a,Chen2022}. 

\subsubsection{About GMMSeg  \cite{Chen2022}}

Some key differences exist with Liang et al \cite{Chen2022}'s formalism. 
First, Liang et al \cite{Chen2022} use fixed uniform priors  $p(c)=\frac{1}{\mathcal{C}}$ for each class while our model is estimating those priors during the training. 
Secondly,  Liang et al \cite{Chen2022} couple the EM optimization for learning GMMs with learning features, while the parameters of the GMMs are learnable parameters of our model. Thirdly the weights of the Gaussian components have been chosen  equiprobable $\omega=\frac{1}{G}$ by  Liang et al \cite{Chen2022} (to ease estimation with the EM algorithm and  forcing it to split the $N_c$ training samples evenly across the $G$ Gaussians), but these weights are  estimated instead in our model.

\subsubsection{Comparison to SDGM \cite{hayashi2021a}}

Instead of using means and covariances as parameters of Gaussians, SDGM \cite{hayashi2021a} uses a tensor $\mathbf{w}$ as parameter of a multivariate Gaussian of the form $\propto \exp \left( \mathbf{w}^T\phi \right)$ where tensor $\phi$ captures a second order polynomial form  of the input feature. Both $\mathbf{w}$ and $\phi$ are vectors of higher dimension $H=1+D(D+3)/2$ than the feature space dimension $D$, and 
their GMM layer is more akin to a polynomial neural network layer \cite{PolynomialNetworkPAMI2022}.
The Elbo loss is used for optimisation instead of the cross entropy.

\subsubsection{Dimensionality reduction}

In Section \ref{sec:classification:results}, we investigate several strategies for reducing the dimension $D$ of the features in the CLIP and ImageBind embedding spaces. One well know strategy is in using a standard learnt linear layer reducing the dimension $D$ to dimension $d$ before the classifier layer. 
Note that for a Softmax  classifier layer, the resulting dimension $d$ needs to be the number of classes  $d=\mathcal{C}$. 
However here using GMM-based classifiers, $d$ can be set as any integer $d\in \mathbb{N}$. In our experiments we mainly test for $d=D$, $d=\mathcal{C}$ and some selected lower dimensions.

\section{Computational complexity DGMMC-S Vs SDGM}
\label{sec:parameters:DGMMC:SDGM}

Table \ref{fig:model_parameters} provides the number of trainable parameters for several of our  tested DGMMC-S classifiers (bottom two rows, \#), and, for comparison, also provides the number    
parameters for  models  SDGM-F  as ratio:
\begin{equation}
\times\text{SDGM-F}(G,d) =\frac{\# \text{SDGM-F}(G,d)}{\# \text{DGMMC-S}(G,d)} 
\label{eq:ratio:SDGM-F}
\end{equation}
and similarly for SDGM-D:
\begin{equation}
 \times\text{SDGM-D}(G,d) = \frac{\# \text{SDGM-D}(G,d)}{\# \text{DGMMC-S}(G,d)} 
 \label{eq:ratio:SDGM-D}
\end{equation}
where  ``\# M" denotes ``number of parameters for classifier M".
From Tab. \ref{fig:model_parameters}, we note:
\begin{itemize}
\item the number of parameters for  SDGM-F is about $d$ times the number of parameters of our model DGMMC-S:
$$
\#\text{SDGM-F} \simeq d \times  \#\text{DGMMC-S}
$$
\item For SDGM-D, we note its number of parameters about $4$ times the number of parameters of our model DGMMC-S:
$$
\#\text{SDGM-D} \simeq 4 \times  \#\text{DGMMC-S}
$$
\end{itemize}

\section{Classification accuracy on CLIP and ImageBind}
\label{sec:classification:results}

We assess the classification accuracy of  our classifier  DGMMC-S in comparison to   SDGM-F Vs SDGM-D classifiers, with or without linear layer.

\paragraph{Datasets}
For this experience we are considering the MNIST ($C=10$), CIFAR10 ($C=10$), CIFAR100 ($C=100$)  and ImageNet ($C=1000$) 
  datasets \cite{MNISTDataset,Krizhevsky09,ImageNet-Dataset} for the task of image classification with both CLIP and ImageBind as feature extractor.
The ESC-50 ($C=50$) dataset \cite{piczak2015dataset} is also tested using ImageBind as the feature extractor.

\paragraph{Embedded spaces}
For all datasets we compute their CLIP  features \cite{CLIP2021} (except audio dataset ESC-50) and ImageBind features \cite{Girdhar_2023_CVPR} (all datasets).
Unlike CLIP, ImageBind is able to compute features for other modalities (including audio) than images, hence   classification accuracy results are also reported with the audio dataset ESC-50 ($C=50$) with ImageBind features only.
Both CLIP and ImageBind  have been provided with their trained weights and all datasets used in this paper have their  CLIP/ImageBind features pre-computed and saved locally for efficiency. CLIP provides $D=768$ features and ImageBind provides  $D=1024$ features. 
When a linear layer is used before the classifier to reduce the dimension $D$ to $d$, several dimensions $d$ of feature spaces are tested.

\paragraph{Experiments} Acting as a baseline with $G=1$, the first experiment (Sec. \ref{sec:baseline:accuracy:no:linear:transformation})  performs classification directly on the embedding feature spaces without linear transformation (or equivalently the linear transformation matrix is set to the identity matrix). 
The second experiment (Sec. \ref{sec:learnt:W}) learns a linear layer in addition to the  parameters of the classifiers, while the last experiment proposes to use principal components computed with PCA instead of learning a linear layer (Sec. \ref{sec:PCA}).

\subsection{Classification on embedded feature spaces with $G=1$}
\label{sec:baseline:accuracy:no:linear:transformation}

DGMMC-S,    SDGM-F and SDGM-D classifier performances are compared here in the simpler case where only one $G=1$ Gaussian is used to represent each class.

\paragraph{Pipeline summary}
The data is transformed as follow:
\begin{equation}
  \text{Data} \rightarrow \underbrace{\text{feature: } \mathbf{x}\in \mathbb{R}^D}_{\text{CLIP or ImageBind}} \rightarrow 
\mathbf{x}=\mathbf{y}\rightarrow \underbrace{\text{Classifier}(\mathbf{y}) }_{\text{learnt }}  
\label{eq:pipeline:1}
\end{equation}

\paragraph{Training Protocol} All classifiers, DGMMC-S,    SDGM-F and SDGM-D, are trained using  SGD optimizer with momentum (set to $0.9$) and Nesterov acceleration. The initial learning rate is set at $1e-3$ and decreases to $1e-4$ using a cosine annealing scheduler over 30 epochs.

\paragraph{Results}
Table \ref{tab:pretrained_results} shows the classification mean accuracy with standard deviation (computed over 3 runs) obtained with  DGMMC-S,    SDGM-F and SDGM-D.
First, ImageBind features perform best on all datasets in comparison to CLIP features. With ImageBind, our proposed classifier  DGMMC-S performs the best in comparison to  SDGM-F and SDGM-D that use more complex covariance matrices (at the exception of audio dataset ESC-50 where SDGM-D performs best). On the other hand, on CLIP feature space, SDGM classifiers perform better than the simpler DGMMC-S. Note that
SDGM-F could not be run for ImageNet  for the machine used for our experiments: this classifier has too many parameters (SDGM-F has about $D$ times more parameters than DGMMC-S, c.f. Tab. \ref{fig:model_parameters} with  $G=1$, $C=1000$, $G=1$ and $d=D=768$ (CLIP) and $d=D=1024$ (ImageBind)).

\begin{table}[!h]
    \centering
    \resizebox{\linewidth}{!}{
    \begin{tabular}{|c|c|c|c|}
    \hline
     &\multicolumn{3}{c|}{CLIP ($d=D=768$)}\\
     \hline
        Datasets & SDGM-F & SDGM-D & DGMMC-S \\
        \hline
        MNIST & \bf{98.9} (0.1) & \underline{98.1} (0.1) & 86.7 (0.6) \\
        \hline
        CIFAR10 & \bf{97.8} (0.003) & \underline{97.5} (0.04) & 88.7 (2.9) \\
        \hline
        CIFAR100 & \bf{85.9} (0.2) & \underline{84.1} (0.3) & 72.1 (1.0) \\
        \hline
        ImageNet & x & \bf{80.1} & 71.6  \\
        \hline
        \multicolumn{4}{c}{}\\
        \hline
        &\multicolumn{3}{c|}{ImageBind ($d=D=1024$)}\\
        \hline
        Datasets & SDGM-F & SDGM-D & DGMMC-S \\
        \hline
        MNIST & \underline{94.4} (1e-3) & 93.4 (0.1) & \bf{99.3} (0.1) \\
        \hline
        CIFAR10 & \underline{98.3} (0.01) & 98.1 (0.02) & \bf{98.7} (4e-3) \\
        \hline
        CIFAR100 & \underline{82.4} (0.2) & 79.5 (0.6) & \bf{91.2} (0.02) \\
        \hline
        ImageNet & x & 59.8 & \bf{84.1}  \\
        \hline\hline
        ESC-50 & \underline{81} (2) & \bf{82.3} (2.3) & 57.3 (8.9) \\
        \hline
    \end{tabular}
    }
    \caption{Accuracy \% (mean and standard deviation computed over 3 runs, excluding ImageNet) for classification with GMM based classifiers ($G=1$)  (cf. Sec. \ref{sec:baseline:accuracy:no:linear:transformation}, pipeline (\ref{eq:pipeline:1})). Bold highlight  best accuracy with  underline for second best  accuracy.}
  
    \label{tab:pretrained_results}
\end{table}

\subsection{Classifiers with a learnt transformation}
\label{sec:learnt:W}

DGMMC-S,    SDGM-F and SDGM-D classifier performances are compared here with using a learnt linear transformation for reducing the dimensionality. Additionally  we test both $G=1$ and $G=2$ Gaussian components as part of the GMMs representing   each class.
In this experience, we are considering a linear layer with bias in order to change the $D$-dimensional feature space into a $d$-dimensional feature space.
This operation allows to reduce the number of features used by the classifiers, hence limiting of the number of trainable parameters to learn during the training even if we need to consider the parameters of the linear layer that now also need to be trained.

\paragraph{Pipeline summary} The data is transformed as follow:
\begin{equation}
  \text{Data} \rightarrow \underbrace{\text{feature: } \mathbf{x}\in \mathbb{R}^D}_{\text{CLIP or ImageBind}} \rightarrow\underbrace{ 
 \mathrm{lin}(\mathbf{x})=\mathbf{y}\in \mathbb{R}^d \rightarrow \text{Classifier}(\mathbf{y}) }_{\text{learnt }}  
 \label{eq:pipeline:2}
 \end{equation}

\paragraph{Training Protocol} 
The models are trained using and SGD optimizer with momentum (set to $0.9$) and Nesterov acceleration. The initial learning rate is set at $1e-3$ and decreases to $1e-4$ using a cosine annealing scheduler over 50 epochs.

\paragraph{Results}
Table \ref{tab:pretrained_CIFAR100} provides the averaged accuracy results and standard deviation (over 5 runs) on the CIFAR10 and CIFAR100 datasets.
 Choosing $G=2$ Gaussian components sometimes improves slightly classifier's performances (DGMMC-S and SDGM-D): two Gaussians components may be helpful in   compensating the simplicity of one spherical or diagonal Gaussian representing a class. Best performances are again observed for ImageBind feature space in comparison to CLIP.  A higher dimension $d$ allows to improve all results for CIFAR10/100 (Tab. \ref{tab:pretrained_CIFAR100}) and most classifiers accuracies on  ImageNet dataset (Tab. \ref{tab:pretrained_ImageNet}) and audio   dataset ESC-50 (Tab. \ref{tab:pretrained_ESC50}).

\begin{table*}[h]
\begin{subfigure}{0.5\textwidth}
\centering
\begin{tabular}{|c|c|c|c|}
\hline 
Model & G & $d=2$ & $d=10$\\
\hline 
\hline
\multirow{2}{*}{\underline{SDGM-F}} & 1 & 94.9 (0.4)& \bf{97.8} (0.1)\\
 & 2 & 94.4 (0.5) & 97.8 (0.1)\\
\hline
\multirow{2}{*}{SDGM-D} & 1 & 94.7 (0.3) & \bf{97.7} (0.02)\\
& 2 & 94.5 (0.4) & 97.7 (0.02)\\
\hline
\multirow{2}{*}{\textbf{DGMMC-S}} & 1& 91.9 (01.1)& \bf{97.8} (0.1)\\
 & 2& 93.6 (0.3)  & 97.7 (0.1)\\
\hline
\end{tabular}
\caption{\textbf{CIFAR10}-CLIP}
\end{subfigure}
\begin{subfigure}{0.5\textwidth}
\centering
\begin{tabular}{|c|c|c|c|}
\hline 
Model & G & $d=2$ & $d=10$\\
\hline 
\hline
\multirow{2}{*}{\underline{SDGM-F}} & 1 & 94.4 (0.3)  & \bf{98.8} (0.02)\\
 & 2 &  93.9 (0.3) & 98.7 (0.02) \\
\hline
\multirow{2}{*}{SDGM-D} & 1 & 94.5 (0.3) & \bf{98.5} (0.1) \\
 & 2 & 93.8 (0.7) & 98.5 (0.1)\\
\hline
\multirow{2}{*}{\textbf{DGMMC-S}} & 1& 93.6 (1.5)& 98.7 (0.1) \\
 & 2& 95.0 (0.3) & \bf{98.8} (0.05)\\
\hline
\end{tabular}
\caption{\textbf{CIFAR10}-ImageBind*}
\end{subfigure}
\vspace{0.5cm}

\begin{subfigure}{0.5\textwidth}
\centering
\begin{tabular}{|c|c|c|c|}
\hline 
Model & G & $d=32$ & $d=100$\\
\hline 
\hline
\multirow{2}{*}{\bf{SDGM-F}} & 1 & 84.2 (0.4) & \bf{85.7} (0.1)\\
 & 2 & 84.1 (0.2)&  85.6 (0.1)\\
\hline
\multirow{2}{*}{\underline{SDGM-D}} & 1 & 84.2 (0.3)& 85.3 (0.1)\\
 & 2 &  84.2 (0.2)& \bf{85.5} (0.3)\\
\hline
\multirow{2}{*}{DGMMC-S} & 1& 81.5 (0.1) & \bf{84.3} (0.1)\\
 & 2&  81.5 (0.1) & 84.1 (0.3)\\
\hline
\end{tabular}
\caption{\textbf{CIFAR100}-CLIP}
\end{subfigure}
\begin{subfigure}{0.5\textwidth}
\centering
\begin{tabular}{|c|c|c|c|}
\hline 
Model & G & $d=32$ & $d=100$\\
\hline 
\hline
\multirow{2}{*}{\bf{SDGM-F}} & 1 & 89.0 (0.1) & \bf{90.5} (0.1)\\
 & 2 &  88.7 (0.1) & 90.2 (0.1)\\
\hline
\multirow{2}{*}{SDGM-D} & 1 & 81.6 (1.3) & \bf{83.4} (0.6)\\
 & 2 & 73.1 (2.0) & 77.6 (0.3)\\
\hline
\multirow{2}{*}{\underline{DGMMC-S}} & 1& 86.3 (0.3) & 89.1 (0.5)\\
 & 2& 86.6 (0.3)& \bf{89.7} (0.2)\\
\hline
\end{tabular}
\caption{\textbf{CIFAR100}-ImageBind*}
\end{subfigure}
\caption{Accuracy \% (mean and standard deviation computed over 5 runs) for classification of datasets \textbf{CIFAR10} and \textbf{CIFAR100}  (cf. Sec \ref{sec:learnt:W}, pipeline (\ref{eq:pipeline:2})). Best model in bold with second best underlined (when equal accuracy, the model with the lowest number of parameters is ranked first), and overall best performing features indicated with *.}
\label{tab:pretrained_CIFAR100}
\end{table*}

\begin{table*}[h]
\begin{subfigure}{0.5\textwidth}
\small
\begin{tabular}{|c|c|c|c|c|c|}
\hline 
Model & G & $d=128$ & $d=768$ & $d=1000$ & $d=1024$ \\
\hline 
\hline
\multirow{2}{*}{\underline{SDGM-F}} & 1 & 81.1 & x& x & x\\
 & 2 & \bf{81.3} & x & x  & x\\
\hline
\multirow{2}{*}{\bf{SDGM-D}} & 1 & 82.6 & 83.7& 82.5 & \bf{83.8}\\
 & 2 & 82.6& 83.7& 82.4 & 83.8\\
\hline
\multirow{2}{*}{DGMMC-S} & 1& 69.6& 79.1& 79.9& 79.5\\
 & 2& 69.8 & \bf{80.3}& 80.0& 79.9\\
\hline
\end{tabular}
\caption{\textbf{ImageNet}-CLIP*}
\end{subfigure}
\begin{subfigure}{0.5\textwidth}
\small
\begin{tabular}{|c|c|c|c|c|c|}
\hline 
Model & G & $d=128$ & $d=768$ & $d=1000$ & $d=1024$ \\
\hline 
\hline
\multirow{2}{*}{\bf{SDGM-F}} & 1 & \bf{83.5} & x &x  &x\\
 & 2 & 83.5&x&x &x\\
\hline
\multirow{2}{*}{SDGM-D} & 1 & 80.9 & 81.9 & 82.3 & \bf{82.4}\\
 & 2 & 80.0& 81.6 & 81.9 & 82.0\\
\hline
\multirow{2}{*}{\underline{DGMMC-S}} & 1& 68.8& 81.0& 81.4 & 81.7\\
 & 2& 69.3& 82.6& 83.0& \bf{83.1}\\
\hline
\end{tabular}
\caption{\textbf{ImageNet}-ImageBind}
\end{subfigure}
\caption{Accuracy \%  for classification of dataset \textbf{ImageNet} (cf. Sec \ref{sec:learnt:W}, pipeline (\ref{eq:pipeline:2})). Best model in bold with second best model underlined, and overall best performing features indicated with *.}
\label{tab:pretrained_ImageNet}
\end{table*}

\begin{table}[!h]
\centering
\small
\begin{tabular}{|c|c|c|c|c|c|}
\hline 
Model & G & $d=3$ & $d=50$ & $d=768$ & $d=1024$ \\
\hline
\hline
\multirow{2}{*}{\underline{SDGM-F}} & 1 & 50 (4) & \bf{85} (2) & 85 (2) & 84 (1)\\
 & 2 & 49 (2)& 84 (2) & 84 (3) & 82 (2)\\
 \hline
 \multirow{2}{*}{\bf{SDGM-D}} & 1 & 38 (2) & 82 (2) & 86 (3) & 86 (2) \\
 & 2 & 36 (2)  & 83 (3) & 86 (3) & \bf{87} (3) \\
 \hline
 \multirow{2}{*}{DGMMC-S} & 1&  37 (3) & \bf{81} (3) & 77 (2) &  73 (5)\\
 & 2&  35 (2) & 81 (2) & 80 (6) &  81 (3)\\
 \hline
\end{tabular}
\caption{Accuracy \% (mean and standard deviation computed over 3 runs) for classification of dataset \textbf{ESC-50} with ImageBind features (cf. Sec. \ref{sec:learnt:W}, pipeline (\ref{eq:pipeline:2})). Best model in bold with second best model underlined.}
\label{tab:pretrained_ESC50}
\end{table}

\paragraph{Remarks} Reducing the number of features using a linear layer allows to limit the number of trainable parameters in the classifiers and we observe that  accuracy results with the linear layer compares well in comparison to using features directly provided by the pretrained embedding space. 
However, the linear layer is dependent of the hyperparameter $d$, which need to be set arbitrarily.
We propose next to replace this linear layer by a PCA decomposition in order to reduce the feature space to a suitable $d$-dimensional eigenspace for performing classification.  

\subsection{Using PCA as linear transformation}
\label{sec:PCA}

The linear layer is now replaced by computing the mean and the Principal Components using all training features  (all classes), and retaining the ones with the highest eigenvalues for projection in a reduced $d$-dimensional space. 

\paragraph{Pipeline summary} The data is transformed as follow:
\begin{equation}
  \text{Data} \rightarrow \underbrace{\text{feature: } \mathbf{x}\in\mathbb{R}^D}_{\text{CLIP or ImageBind}} \rightarrow \underbrace{\mathrm{lin}(\mathbf{x})=\mathbf{y}\in \mathbb{R}^d}_{\text{PCA }} \rightarrow \underbrace{\text{Classifier}(\mathbf{y}) }_{\text{learnt}}  
\label{eq:pipeline:PCA}
\end{equation}

\paragraph{Training Protocol} 
We replace the linear layer used in the previously (Sec. \ref{sec:learnt:W}) to reduce the dimension space of the features by a PCA decomposition. 
The cumulative explained variance ratio (Eq. \ref{eq:cumulative_variance_ratio}) can then be used to  select the number $d$ of eigenvectors capturing a reasonable chuck of  the information: 
\begin{equation}
\label{eq:cumulative_variance_ratio}
   \text{variance ratio} (d)= \frac{\sum_{j=1}^{d}\lambda_j}{\sum_{j=1}^{D}\lambda_j}
\end{equation}
where $D$ is the number of dimension in the  embedded feature spaces ($D=768$ for CLIP and $D=1024$ for ImageBind), $d$ represents the first $d$ principal components and $\lambda_j$ is the variance (eigenvalue) of the data around the $j$-th eigenvectors. Classifiers all use $G=1$ Gaussian component per class.

\begin{figure*}[h!]
        \subfloat[]{%
            \includegraphics[width=.48\linewidth]{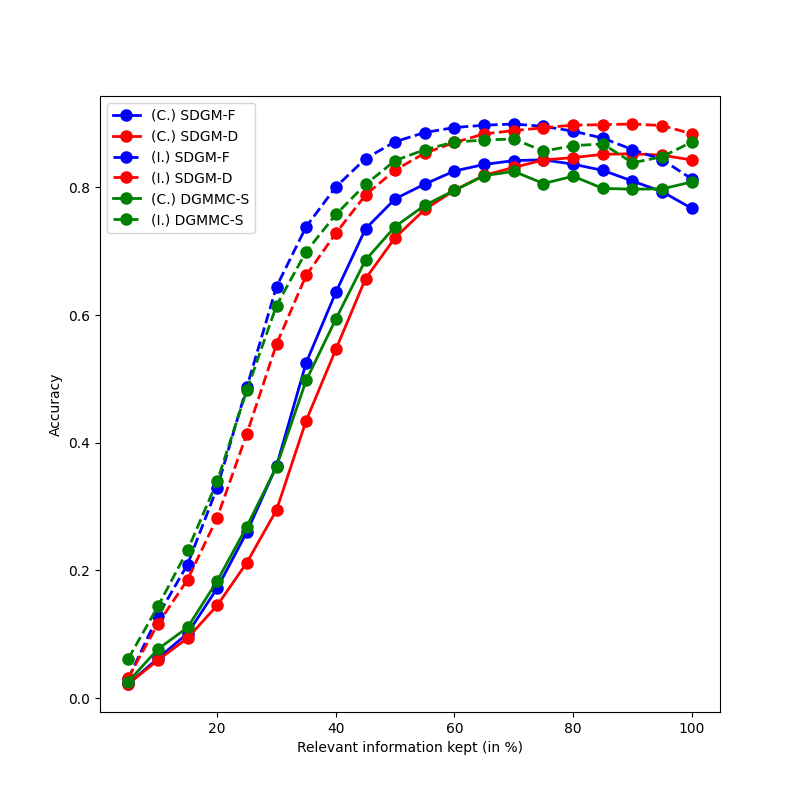}%
        }\hfill
        \subfloat[]{%
            \includegraphics[width=.48\linewidth]{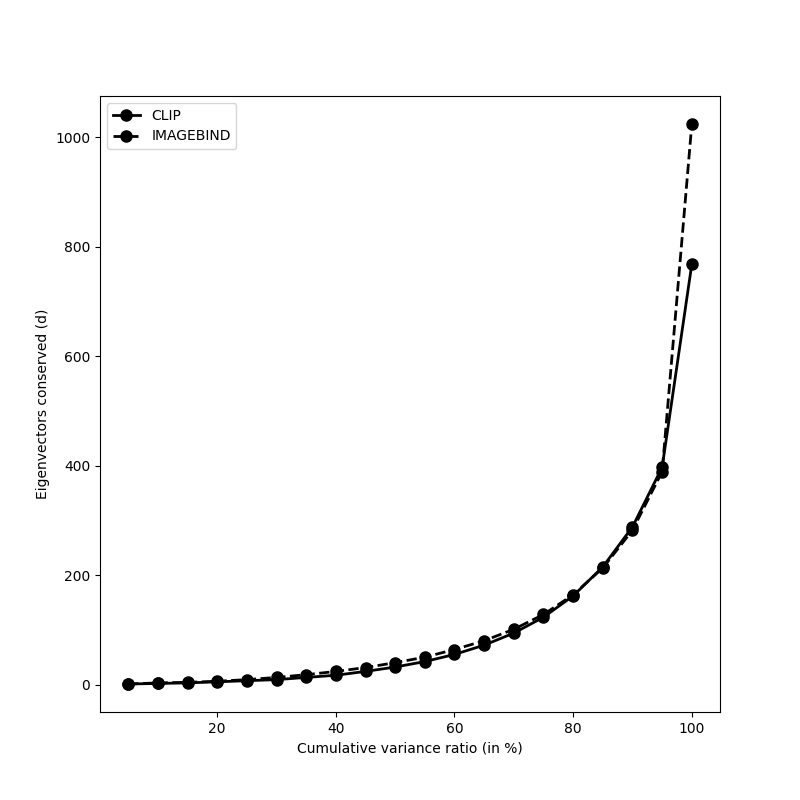}%
        }\\
        \subfloat[]{%
            \includegraphics[width=.48\linewidth]{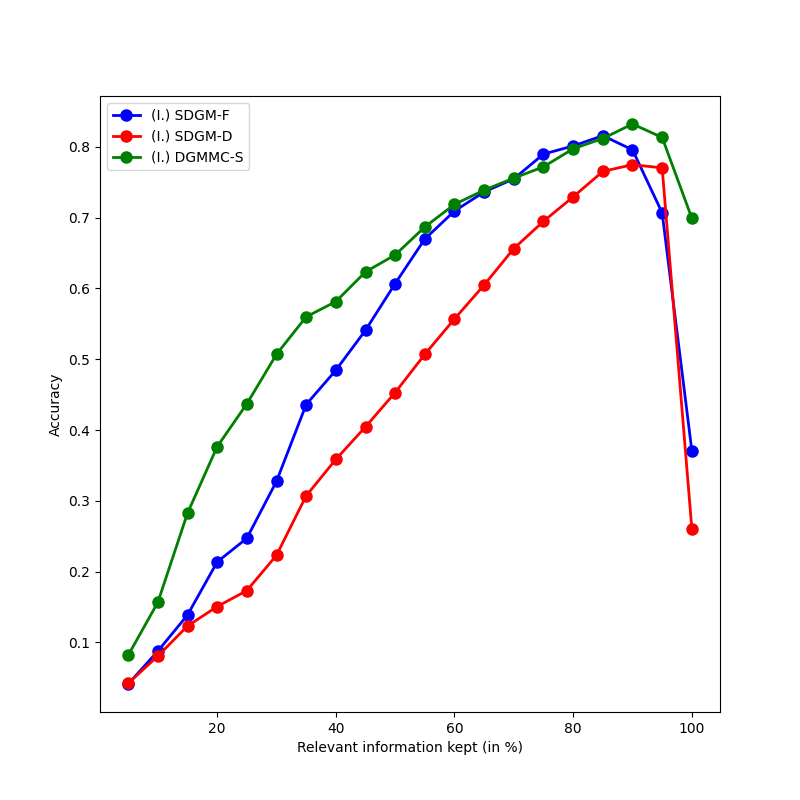}%
        }\hfill
        \subfloat[]{%
            \includegraphics[width=.48\linewidth]{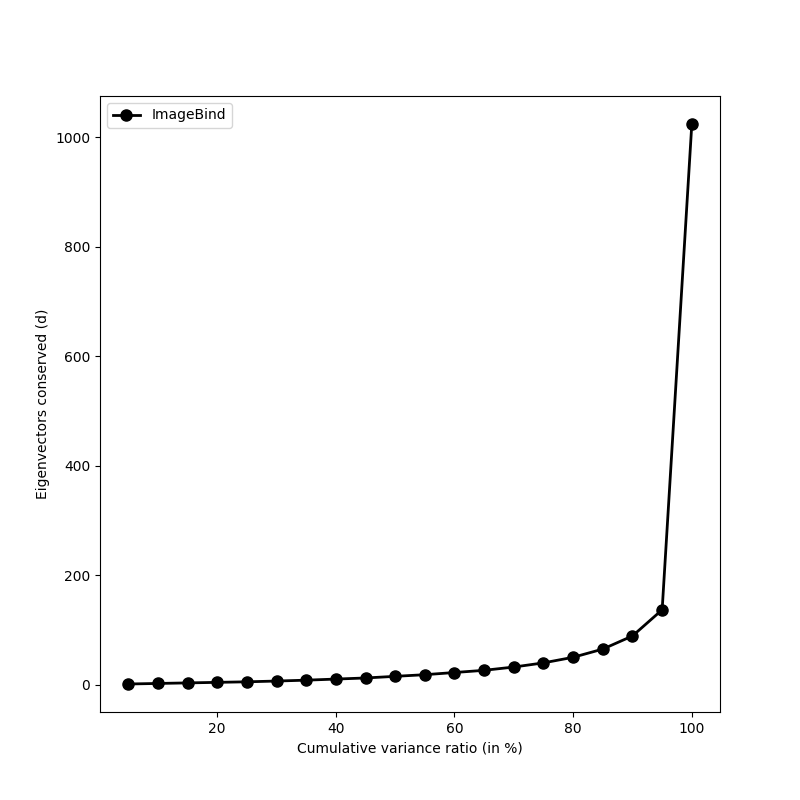}%
        }
        \caption{Evolution of the accuracy depending of the cumulative variance ratio kept (in \%) after PCA decomposition on features provided by the pretrained embedding spaces CLIP (plain lines) and ImageBind (dashed lines). The results are provided for the \textbf{CIFAR100} dataset ((a) and (b)) and the \textbf{ESC-50} dataset ((c) and (d)) where 20 different percentages (from 5\% to 100\%) have been used to plot the curves (cf. Sec. \ref{sec:PCA}).}
        \label{fig:curves_PCA}
    \end{figure*}

\paragraph{Results} 
Fig. \ref{fig:curves_PCA} shows the accuracy of the classifiers with respect to the explained variance for  CIFAR100 (images) dataset and the ESC-50 (audio sounds) datasets. We note that ImageBind features again provide best result in comparison to CLIP features. For ESC-50 we note that accuracy falls when too many principal components are used ($d$ high: principal components associated with lowest eigenvalues may capture noise not representative of the classes and inclusion of these dimensions impact negatively the performance of the classifiers.
Tab. \ref{tab:ImageNet_PCA} shows the accuracy obtained for ImageNet dataset using 80\% of the variance ratio.

\begin{table}[!h]
    \centering
   
    \begin{tabular}{|c|c|c|}
    \hline
        Embedding & CLIP & ImageBind \\
        \hline
        Model &  $d=221$ & $d=220$ \\
        \hline
        SDGM-F & 80.2  & 80.9  \\
        \hline
        SDGM-D & 81.4  & \bf{82.0}  \\
        \hline
        DGMMC-S & 76.9  & 79.3  \\
        \hline
    \end{tabular}
    \caption{Accuracy on \textbf{ImageNet} using  both DGMMC and SDGM classifiers when only using 80\% of the variance ratio resulting in keeping the $d$ most relevant PCA eigenvectors (cf. Sec. \ref{sec:PCA}, pipeline (\ref{eq:pipeline:PCA})).}
    \label{tab:ImageNet_PCA}
\end{table}

\section{Conclusions}

Both ImageBind and CLIP features were trained using contrastive learning that may explain why, in practice, we did not notice major advantages in using more than one Gaussian component in the GMMs encoding the classes. 
Depending on the datasets and the embedding feature spaces, estimating full covariance matrices (with SDGM-F) provide the best results with CLIP, however simpler covariance matrices, diagonal (SDGM-D) and spherical (DGMMC-S) perform better  with ImageBind and with less parameters.
Occasionally having $G=2$ Gaussian components in the GMMs may help DGMMC-S to perform as well as SDGM-F with $G=1$.
ImageBind embedded space has a higher dimensionality $D=1024$ than CLIP $D=768$ and ImageBind  has shown to perform better  than CLIP most of the time.  Decreasing the $D$-dimensional embedded space using a PCA based linear transform to a lower $d-$dimensional one has also shown to be valuable in reducing complexity while maintaining good performance. We have shown that the hyperparameter choice of $d$  can be selected conveniently using the cumulated variance ratio.
We have extended SDGM classifiers by proposing to use spherical covariance matrices for Gaussians as part of our DGMMC-S classifier and all classifiers have been  compared on the several datasets.  
Table \ref{tab:Summary} summarizes the best accuracy results: DGMMC-S has been shown to perform very well in practice on the ImageBind embedding space (without the need of linear layer preceding the classifier), 
and only the audio dataset is better classified when using a learnt linear layer with SDGM-D.

\section*{Acknowledgments}
 This research was conducted with the financial support of
Science Foundation Ireland under Grant Agreement no. 13/RC/2106\_P2 at   ADAPT, the SFI Research Centre for AI-Driven Digital Content Technology, that is funded by Science Foundation Ireland through 
the SFI  Research Centres Programme and is co-funded by the European Regional Development Fund. 
For the purpose of Open Access,
the authors have applied a CC BY public copyright licence to any Author Accepted Manuscript version arising from this submission.


\end{document}